# Data Augmentation of Railway Images for Track Inspection

S. Ritika, Dattaraj Rao

General Electric

## ABSTRACT

Regular maintenance of all the assets is pivotal for proper functioning of railway. Manual maintenance can be very cumbersome and leave room for errors. Track anomalies like vegetation overgrowth, sun kinks affect the track construct and result in unequal load transfer, imbalanced lateral forces on tracks which causes further deterioration of tracks and can ultimately result in derailment of locomotive. Hence there is a need to continuously monitor rail track health.

Track anomalies are rare with the skew as high as one anomaly in millions of good images. We propose a method to build training data that will make our algorithms more robust and help us detect real world track issues. The data augmentation will have a direct effect in making us detect better anomalies and hence improve time for railroads that is spent in manual inspection.

This paper talks about a real-world use-case of detecting railway track defects from a camera mounted on a moving locomotive and tracking their locations. The camera is engineered to withstand the environment factors on a moving train and provide a consistent steady image at around 30 frames per second. An image simulation pipeline of track detection, region of interest selection, augmenting image for anomalies is implemented. Training images are simulated for sun kink and vegetation overgrowth. Inception V3 model pretrained on Imagenet dataset is finetuned for a 2 class classification. For the case of vegetation overgrowth, the model generalizes well on actual vegetation images, though it was trained and validated solely on simulated images which might have different distribution than the actual vegetation. Sun kink classifier can classify professionally simulated sun kink videos with a precision of 97.5%.

## INTRODUCTION

Regular maintenance of all the assets is pivotal for proper functioning of railway. Track is a very important asset for railway. Usually the maintenance is done manually and can be very cumbersome [1]. Also, this leaves room for errors and misses which can cause further deterioration of locomotive as well as track and might ultimately lead to derailment. Track is constructed in a particular way by stacking ties on layers of ballast to ensure the load is transferred equally throughout the construct. But if there is growth of vegetation it can reduce the efficiency of ballast in load transferring and can ultimately affect the track [1]. During extreme climatic changes the track can deform. During summer, due to heat the tracks may expand and if proper clearance is not provided it can get bent. This condition is termed sun kink. Also, if locomotive isn't designed properly, high imbalanced lateral forces on track from locomotive can be the cause of sun kink. This is very detrimental as it causes high vibration in locomotive running on a sun kinked track and can ultimately cause of derailment. This problem has been attempted using conventional image processing technique in the past. [2][3] But these techniques fail to generalize to different environmental and track conditions.

# PROBLEM STATEMENT

Deep neural networks have the constraint that they require a lot of data to be trained without overfitting. Fine tuning a model pre-trained on datasets like Imagenet, COCO, ILSVRC which have millions of labeled images helps a lot to bring down the number of training images required. This is because a lower level features of model which has been trained for classification of a particular type of image can be reused to a completely new image dataset. Even after leveraging pre-trained weights, the minimum number of images required might range from few hundreds to thousands depending on the complexity of the task at hand. Hence the biggest challenge for image classification becomes availability of trainable data.

For the task of railway track health monitoring, we have a lot of good track data available. But anomalies like vegetation overgrowth, sun kink are rare and difficult to find. These can be generated manually using tools like paint but it can be a very cumbersome, labor-intensive process. Hence if synthetic data can be generated for the anomalies mentioned above, it can ease the training process and reduce the problem of overfitting.

# APPROACH

The problem can be divided into two parts: finding the track and augmenting it to add anomalies in the region of interest.

## Detecting Track

Track is a very prominent feature in the image since it is a straight converging line. With the placement of camera as seen in figures below, the track position doesn't vary a lot. Hence, a region of interest (ROI) can be demarcated where there is a high possibility of track to be present. Once the ROI is clipped, the exact position of track needs to be found. Since track are straight lines, edge detection can be used for the same. Canny edge detection [4] is found to give very good results once the thresholds are tuned properly. Image can be filtered before edge detection to remove noise. Edge detection results in a cluster of number of lines. We need to extract the tracks out of it. Hough's line transform [5] can be used for the same. Thus, the number of lines have reduced but we didn't get the two lines corresponding to the track yet. The feature that helps at this juncture is that tracks converge. Hence all the lines can be grouped depending on whether their slope is positive or negative. The zero slope lines (mostly due to the track tie) can be ignored in this process. This gives the coordinates of the two lines corresponding to the track. The ROI can be specified utilizing these points.

## Simulating Vegetation Overgrowth

Once the ROI is selected, vegetation needs to be simulated. Vegetation can disrupt the balance of the ballast and can be detrimental to the track. The level of vegetation can be anything from sparse to high. The region can be inside the tracks and over the rails of the track also. Vegetation is usually a cluster of many small grass blobs. Hence the simulation can be generated in the same way. To generate the grass blob greenish pixels can be placed at random places iteratively. Another efficient way for doing this will be using gaussian distribution. Normal, skewed distribution can also be experimented. Random number of such blobs can be placed in the ROI. The number of blobs can control the level of vegetation.

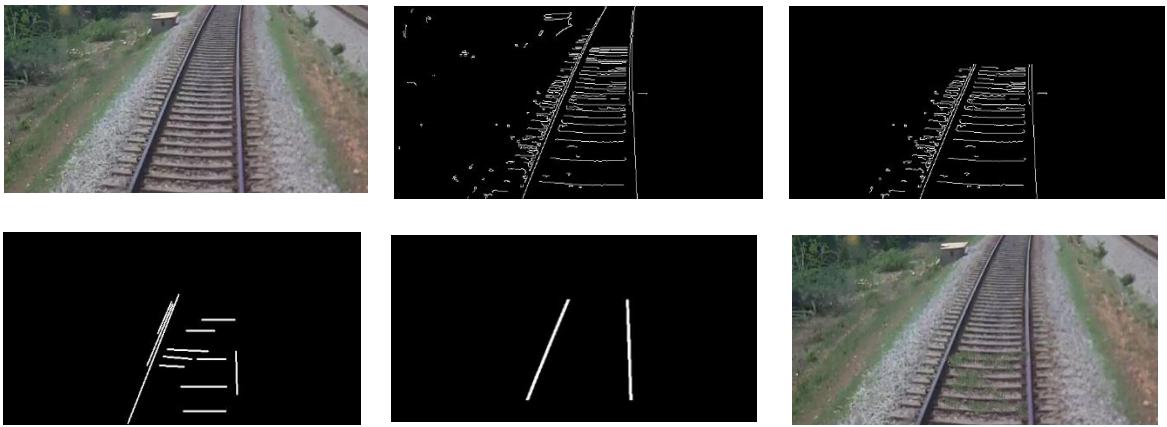

*Figure 1 Step by step process for simulating vegetation overgrowth*

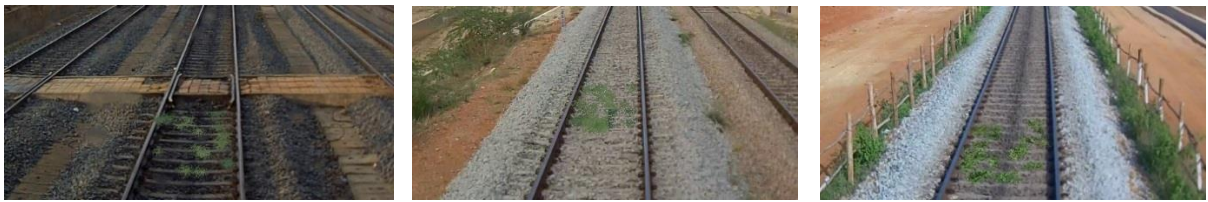

*Figure 2 Examples of images simulated with vegetation overgrowth*

## Simulating Sun kink

Once the track coordinates are detected, the task can be divided into two parts: removing the track and adding a kinked track. To remove the track, a part of ROI can be copied and shifted sideways. This will ensure that the ties look complete. This step can also be used to generate missing rail condition. To generate broken rail condition, two random points on the track can be connected with a dark line. To generate the kinked track, cubic splines can be utilized. Depending on the size of kink, two random points can be chosen on the track and a spline can be drawn utilizing these.

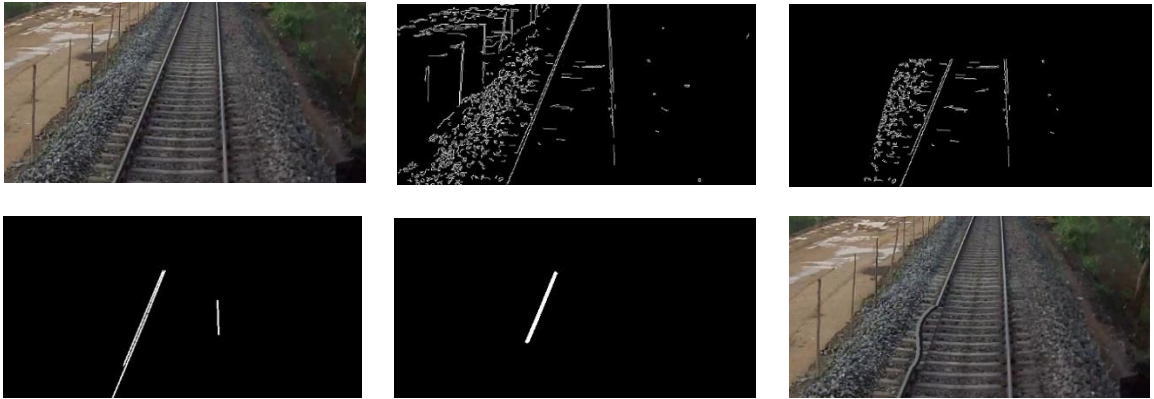

*Figure 3 Step by step process for simulating sun kink*

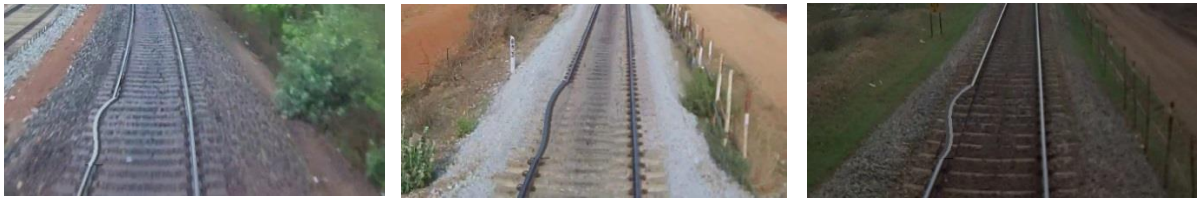

*Figure 4 Examples of images simulated for sun kinks*

# EXPERIMENTAL RESULTS

The experiments were done on the video feed from a front facing camera mounted on the locomotive. An ROI 600*300 was extracted from the image. The image was filtered using gaussian filter. Canny edge detection was applied followed by Hough's line transform. The results for these steps are shown in Figure 1 Step by step process for simulating vegetation overgrowthFigure 1 and Figure 3 above.

## Vegetation Overgrowth

A balanced dataset of 500 images were created comprising of images with vegetation generated by the process described above and healthy track images. The dataset was augmented by varying the brightness, flipping the image horizontally. Inception [6] V3 network pretrained on Imagenet dataset was taken as baseline. The network was fit on the training data by finetuning the last classification layer to obtain a two class classifier. The performance of the model on the test data can be seen in Table 1. An important point to note is that the validation set while training the data was solely the simulated image while the model was tested on real vegetation images. The model was seen to give a good performance even though validation and test data might have different distributions.

*Table 1 Test results on actual vegetation images by Inception V3 model trained solely on images simulated for vegetation overgrowth*

| Actual\Predicted | Positive | Negative |
|---|---|---|
| Positive | 10 | 4 |
| Negative | 1 | 5 |

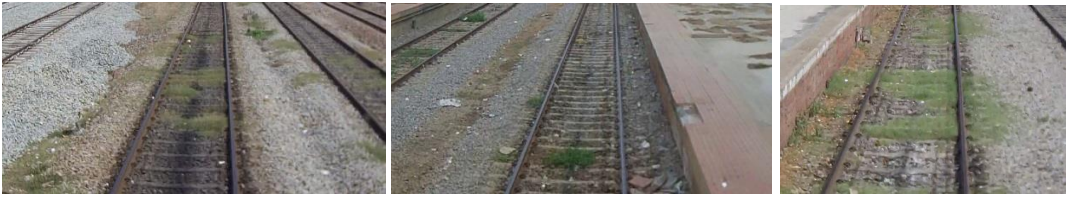

*Figure 5 Vegetation overgrowth predicted correctly by the model*

## Sun kink

A balanced dataset of 500 images was created for a 2-class classification using Inception V3 network similar to the approach taken for vegetation overgrowth. The network was finetuned on the dataset with augmentation of flipping and random brightness adjustments. The trained model was tested on track video professionally simulated for sun kink.

*Table 2 Results on professionally simulated videos of the Inception V3 model trained solely on simulated sun kink images*

| Actual\Predicted | Positive | Negative |
|---|---|---|
| Positive | 6 | 0 |
| Negative | 1 | 5 |

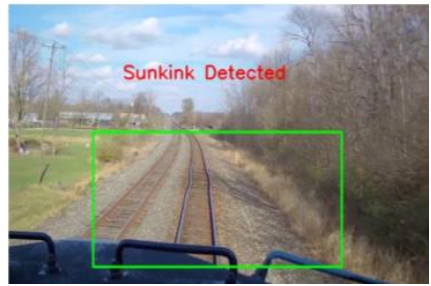

*Figure 6 Sun kink detected correctly on professionally simulated video*

# CONCLUSION

For the domain of rail data, training image generation for anomalies like vegetation and sun kink was attempted using conventional image processing techniques. It was observed that using pre-trained models and data augmentation brings down the number of training images required. The image simulation pipeline used was: track detection, region of interest selection, augmenting image for anomalies. A balanced dataset of 500 images containing healthy track as well as anomaly was generated. Two separate models were trained for vegetation overgrowth and sun kink. Inception V3 model pretrained on Imagenet dataset was finetuned for 2 class classification.  For the case of vegetation overgrowth, the model generalized well on actual vegetation images, though it was trained and validated on simulated images which might have different distribution than the actual vegetation. Sun kink classifier could classify professionally simulated sun kink videos with a precision of 97.5%.